# Salient Slices: Improved Neural Network Training and Performance with Image Entropy

Steven J. Frank and Andrea M. Frank

Abstract—As a training and analysis strategy for convolutional neural networks (CNNs), we slice images into tiled segments and use, for training and prediction, segments that both satisfy an information criterion and contain sufficient content to support classification. In particular, we utilize image entropy as the information criterion. This ensures that each tile carries as much information diversity as the original image, and for many applications serves as an indicator of usefulness in classification. To make predictions, a probability aggregation framework is applied to probabilities assigned by the CNN to the input image tiles. This technique facilitates the use of large, high-resolution images that would be impractical to analyze unmodified; provides data augmentation for training, which is particularly valuable when image availability is limited; and the ensemble nature of the input for prediction enhances its accuracy.

## I.    INTRODUCTION

Effective training of neural networks, particularly convolutional neural networks (CNNs), depends critically on the availability of high-quality data.  While a robust architecture and adequate computational capacity (not to mention patience) are essential, a dearth of data can doom otherwise promising efforts.  Goodfellow et al., 2016 estimates that achieving human-level classification performance requires training a neural network with about 5000 labels per class.  As a result, various strategies for data augmentation and repurposing have been developed to do more with less, and heroic efforts have been made to hoard quality data into vast and expanding repositories for public use.

Sometimes, however, the data simply isn't there.  In our particular area of interest, artwork attribution and the detection of forgeries, the scope of the dataset cannot exceed a particular artist's output.  But many other areas of interest have similar limitations — histology slides may be rare for unusual medical conditions, for example, and legal (e.g., patient privacy) restrictions can prevent widespread dissemination of imaging data that does exist.

A second challenge is image size.  Convolution is a computationally expensive operation; the number of convolutions performed (particularly by the first CNN layers) scales with the product of the resolution dimensions.  Training will be slow for large images, and the input and memory limits of a CNN architecture may be tested.  Conventional strategies of downsampling the image or simplifying the architecture sacrifice the ability of a CNN to classify based on fine or subtle features.

## II.    DOMAIN CHALLENGES

Analyzing artwork for attribution or authenticity based solely on the visual characteristics of the work itself represents a demanding classification task.  Whereas physico-chemical analysis of canvas and paint can scientifically refute authenticity (e.g., due to inconsistencies between the analyzed work and materials available to the artist), art connoisseurs and historians often disagree over the stroke-level and stylistic features that characterize an artist's true work. No accepted methodology exists for identifying such features, which in any case can evolve over an artist's career. A century ago, Rembrandt's total output was estimated at 711 works. That number began to shrink, soon quite dramatically, following the establishment of the Rembrandt Research Project in 1968. Members of this committee, Dutch art historians charged with the task of de-attributing dubious Rembrandts, often disagreed over stylistic criteria, and the very existence of such disagreement frequently resulted in de-attribution. Dozens of works were rejected. By 1989, only 250 works had survived their judgment, although the committee restored 90 or so paintings to the canon before disbanding in 2011.

With so much vigorous and evolving expert disagreement over a single artist's work, the goal of identifying artist-specific (much less universal) indicia of authenticity seems hopeless — making the domain in many ways ideal for analysis with CNNs, which are adaptive and learn their own visual criteria. Numerous researchers (Lecoutre et al., 2017; Balakrishnan et al., 2017; Tan et al., 2016; Saleh et al., 2015; Bar et al., 2014) have used CNNs to categorize art images by movement and style (in some ways a more formidable, if not intractable problem, since art styles resist formal definition and often overlap). Another effort, dubbed PigeoNET, attempted to classify a large collection of paintings by artist rather than style (van Noord et al., 2015). These efforts have all involved whole-image classification and suffer from low image resolutions.

We are aware of very few efforts to perform artist attribution using a trained CNN, and these have involved hand-engineered features (Elgammal et al., 2017; Saleh et al., 2015; Karayev et al., 2014, Johnson et al., 2008; Li et al., 2004; Lyu et al., 2004; Sablatnig et al., 1998).

III.    METHODOLOGY

Here we attempt to perform CNN classification of artwork at high resolution, permitting the computational analysis to consider fine, brushstroke-level detail not visible in coarse images and to avoid the need for subjective, artist-specific feature engineering.

A.  *Image Entropy*

Image fragments, rather than whole images, have been employed in tasks ranging from object detection and classification (Ullman et al., 2001) to identification of lung carcinomas in histopathology images (Yu et al., 2019). Our concern is identifying image fragments most likely to speed CNN training and support testing, i.e., which represent the most visually salient portions of an image for the ultimate classification task.

The entropy of an image, from the purview of information theory, represents the degree of randomness of its pixel values, just as the entropy of a message denotes (as a base-2 log) the amount of useful, nonredundant information that the message encodes:

$$H = -\sum_k p_k \log_2(p_k)$$

In a linear message, $p_k$ is the probability associated with each possible data value $k$. For a two-dimensional eight-bit grayscale image (or one channel of a color image), $k$ spans the 256 possible pixel values [0..255], and the probabilities for an image $x = (x_1, \dots, x_{256})$ are given by:

$$p_k(x) \equiv \frac{\#\{x_i | x_i = k\}}{256},$$

where *#S* denotes the cardinal of set *S*. The entropy can be calculated at each pixel position (*i,j*) across the image. Coded in Python, the operation takes 2-3 sec for an average-sized painting using Google Colabs (although actually generating tiles can take up to 5 min).

For our purposes, we view image entropy as roughly corresponding to the diversity of visual information present in an image. As indicated in Fig. 1, the complexity of an image corresponds intuitively with the entropy and reflects its logarithmic character. To the extent that increasing image entropy correlates with increasingly rich feature content captured in the convolutional layers of a CNN, it provides a useful basis for selecting image tiles. Our approach is to divide an image into discrete but partially overlapping tiles and retain only those tiles whose entropies equal or exceed the entropy of the whole image. Although no subimage will contain as much information *content* as the original, our conjecture is that a subimage with comparable information *diversity* will, on average, be classified like the source image by a trained CNN.

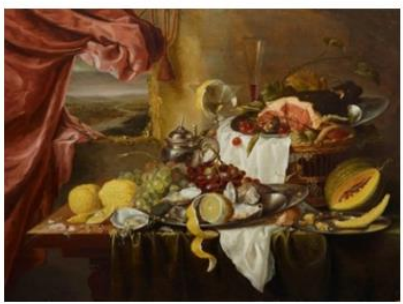 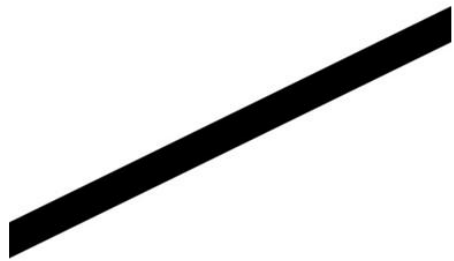 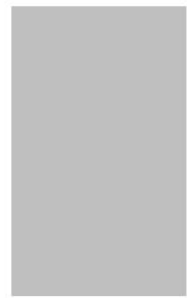

Entropy = 6.98                    Entropy = 0.83                    Entropy = 0.0

Fig. 1: Entropies of a Dutch still life, a diagonal bar, and a uniform field

Fig. 2 suggests the plausibility of this conjecture for representational art. For illustrative purposes, we applied the entropy criterion to 100×100 pixel tiles derived from a source image at relatively low (463×600 pixel) image resolution. The retained subimages capture the most visually complex elements of a portrait — the hands, facial details, elaborate clothing features. They're the parts that attract our information-hungry eyes and, we find, drive the operation of a CNN.

Because information diversity is not the same as information content, however, image entropy does not provide a useful basis for defining a minimum tile size; even very small tiles can have high entropies. We therefore test a stepped sequence of tile sizes to determine the optimal size for our task, and as discussed below, the optimal size depends strongly on the artist studied.

*B. Classification Metrics*

We tested two classification metrics: average tile probability and majority vote. At the tile level, a probability approaching 1 corresponds to a high likelihood that the tile is attributable to the artist under study, while a probability approaching 0 represents a high likelihood that it is not. Assigning an image classification based on average tile probabilities or by majority vote nearly always produces the same result, and when the outcomes differ, there is no evident bias toward one classification. For consistency, we report results in terms of average tile probability.

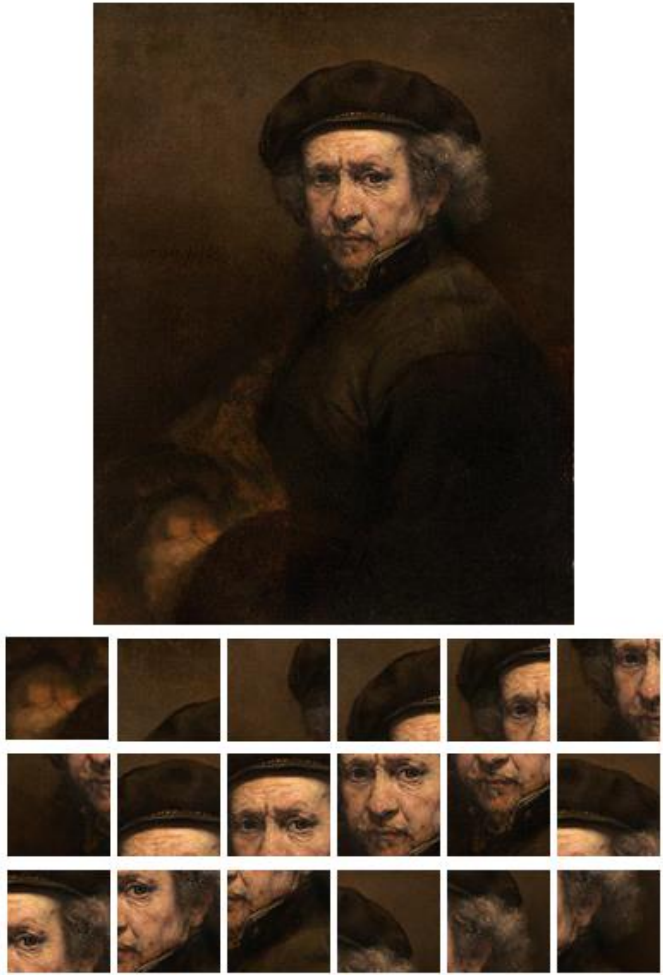

Fig. 2: A source Rembrandt image at relatively low resolution and 100×100 pixel image tiles derived from it using our algorithm

*C. System Design and Datasets*

We investigated whether a CNN could accurately distinguish the work of two artists, Rembrandt and Vincent van Gogh, from their imitators and those whose work they influenced. We focused on Rembrandt's portraits and van Gogh's landscapes. As detailed below, we compared the performance of our algorithm against tile sets selected at random, and also compared square tiles with rectangular tiles whose aspect ratios matched those of the paintings from which they were drawn. The steps of our procedure are summarized in Fig. 3.

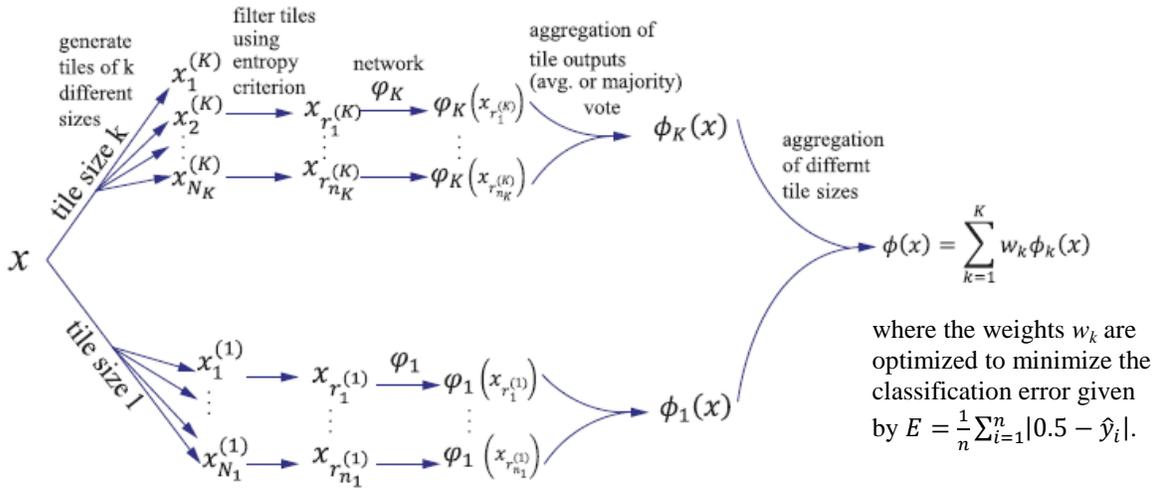

Fig. 3: Summary of classification procedure. A resized image $x$ is decomposed into tiles of $K$ sizes. For each tile size $k \in \{1 \ldots K\}$, the number $N_k$ of tiles $x^{(k)}$ produced from an image will necessarily differ. The tiles are sifted using the image-entropy criterion, yielding, for each tile size $k$, a reduced tile set $x_{r_{n_k}^{(k)}}$ where $n_k < N_k$. The CNN, trained using the Adam optimizer, assigns a classification probability to each tile to produce, for each reduced tile set, a probability vector $\varphi_k$. The probability vectors are aggregated by averaging or majority vote into an output image-level probability $\phi_k(x)$. Finally, the output probabilities from different tile sizes may be combined in a weighted fashion to produce a final output classification $\phi(x)$.

Good performance for our classification task appears to require a simple model. We trained and tested three-layer and five-layer models as well as the VGG16 and ResNet50 architectures, the latter two models pre-trained on the ImageNet dataset. Both of the more complex architectures produced inferior results for van Gogh and failed altogether for Rembrandt, with VGG16 attributing every tile to Rembrandt with a probability of 1.0 and ResNet50 assigning nearly all probabilities within 1% of the decision boundary — i.e., learning failed, even when we varied the learning rate. Further testing demonstrated the superiority of a model with five convolutional layers over a three-layer model, and we found the best learning rate to be 0.001. Three dropout layers proved optimal. Our models used five max pooling layers, five batch normalization layers, sigmoid activation, a binary cross-entropy loss function, and an Adam optimizer. Source code for all tested models has been posted.[1]

Training and use of a CNN with image fragments necessitates relative consistency among digitized image resolutions, i.e., image pixels per unit width or height of canvas in the case of paintings.  Candidate images were therefore screened to ensure that the image resolutions did not vary by more than a factor of about five (to avoid excessive, and potentially distortive, downsampling) but exceeded our target image resolution (to avoid upsampling); specifically, all images were resized to a consistent image resolution of 26.81 pixels per cm of canvas in the case of Rembrandt and 25.00 pixels/cm for van Gogh; the rescaled Rembrandt images had an average size (width × height) of about 2150×2700 pixels and the average size of the rescaled van Goghs was 1780×1440.[2]  The bit depth was eight for all images and tiles.

The choice of Rembrandt portraits and van Gogh landscapes as subjects reflects an effort to generalize as much as possible across representational artists and genres.  Unlike Rembrandt, van Gogh is known for his daring, expressive brushstrokes and dramatic juxtapositions of color — implying that the features important for classifying van Gogh will occur at smaller scales than for Rembrandt.  Our results suggest that a wide range of representational art should fall within the categories broadly delineated by these selected artists and genres.

The hoped-for ability of the system we describe to detect forgeries also points to risks associated with training and test datasets.  John Rewald, a distinguished historian of Impressionism and Post-Impressionism, remarked that van Gogh may well have been forged "more frequently than any other modern master" (Bailey, 1997).  For Rembrandt, the risks may be even worse, since, unlike van Gogh, he had 100 recorded students — and as one scholar has noted, "It was the pupil's business to look like the master" (Brenson, 1985, p. C13).  Can the ground truth of a dataset be trusted in the face of such questions surrounding authenticity?

We have confidence in our datasets thanks to the age and fame of the paintings we analyzed and the persistent scholarly scrutiny to which they have been subjected.  All of the Rembrandt works in our dataset have been authenticated by the Rembrandt Research Project. The works of van Gogh have also undergone extensive analysis, with the most recent catalogue raisonné raising new doubts over 45 paintings and drawings (Hulsker, 1996); once again, the works in our dataset have not been questioned.

Because our technique deliberately avoids feature engineering, success depends on strategies for data curation.  Similarity judgments are necessarily subjective, but we made efforts to span, for our comparative images, a range extending from very close to evocative but readily distinguishable in order to retain fine distinctions without overfitting — that is, to broaden classification capability beyond the specific artists represented in the training set.  We used a split of Rembrandt portraits and comparative portraits selected for varying degrees of pictorial similarity to the Rembrandts. Our van Gogh dataset was evenly divided between his landscapes and those of close imitators, such as the Swiss painter Cuno Amiet, as well as contemporary and later works that clearly show van Gogh's influence.[3] We kept our Rembrandt dataset smaller than the van Gogh set — 51 training/25 test images for Rembrandt, 100 training/52 test images for van Gogh — partly due to the paucity of available images for rigorously authenticated Rembrandt portraits but also with the aim of assessing performance with relatively few training images; only 36 authentic paintings by Rembrandt's compatriot Johannes Vermeer have been identified, for example, so the need for large numbers of training examples would potentially exclude many interesting studies.

We first attempted to identify an optimal CNN architecture. For both artists, we prepared tile sets ranging in size from 100×100 to 650×650 pixels in 50-pixel steps. Above this tile size, we encountered memory errors even with 32 GB of volatile storage. The more serious problem, however, is the diminishing number of generated tiles and the still smaller number that passed the entropy criterion. While the training set can be expanded, a tested image can only yield a certain number of large tiles and only some of these will satisfy the entropy discriminator; as the number of tested tiles decreases, the accuracy-enhancing benefits of ensemble averaging erode.

### D. Tile Size and Selection

The number of training and testing tiles necessarily depended strongly on the tile size. With 50% overlap, we obtained 11,761 100×100 Rembrandt and non-Rembrandt training tiles but only 276 800×800 training tiles. We found that 15-20% of the candidate tiles satisfied the entropy criterion across all tile sizes. For the larger van Gogh training set, we obtained 10,700 100×100 training tiles and 245 800×800 training tiles at 50% overlap. This reflected a much lower retention rate following application of the entropy discriminator, which we discuss below.

### IV. RESULTS AND DISCUSSION

Our initial results, shown in Table 1, demonstrate a pronounced difference between the two artists in terms of optimal tile size. For Rembrandt, classification accuracy peaked at the 450×450 tile size corresponding, roughly, to a face-size feature as illustrated in Fig. 4. For comparative purposes we trained and tested models at a larger resolution (50 pixels per canvas cm), and found that results were substantially worse across all tile sizes — not surprising, because a larger resolution magnifies the image and consequently emphasizes smaller features. Moreover, at this higher resolution, photographic imperfections (blur) became evident.

Peak accuracy for van Gogh, by contrast, occurred at much smaller tile sizes that restrict CNN analysis to finer levels of detail. It appears that, whereas van Gogh's unique signature — the distinctive features that facilitate accurate convolution-based classification — appears at the brushstroke level, what sets Rembrandt apart from his imitators emerges at the level of composition. These results challenge the assumption, common among art connoisseurs, that an artist's signature is invariably found in his or her stroke (van Dantzig, 1973; Laurie, 1932). Perhaps Rembrandt's brushstrokes were rather ordinary, or maybe as a teacher he imparted his mechanics well; perhaps van Gogh's emotive engagement with the canvas resists facile duplication. Such speculations are also facile. What is quite likely, however, is that different representational artists will be most successfully distinguished at scales corresponding to what, in some sense, characterizes their visual uniqueness — and that Rembrandt and van Gogh exemplify opposite poles.

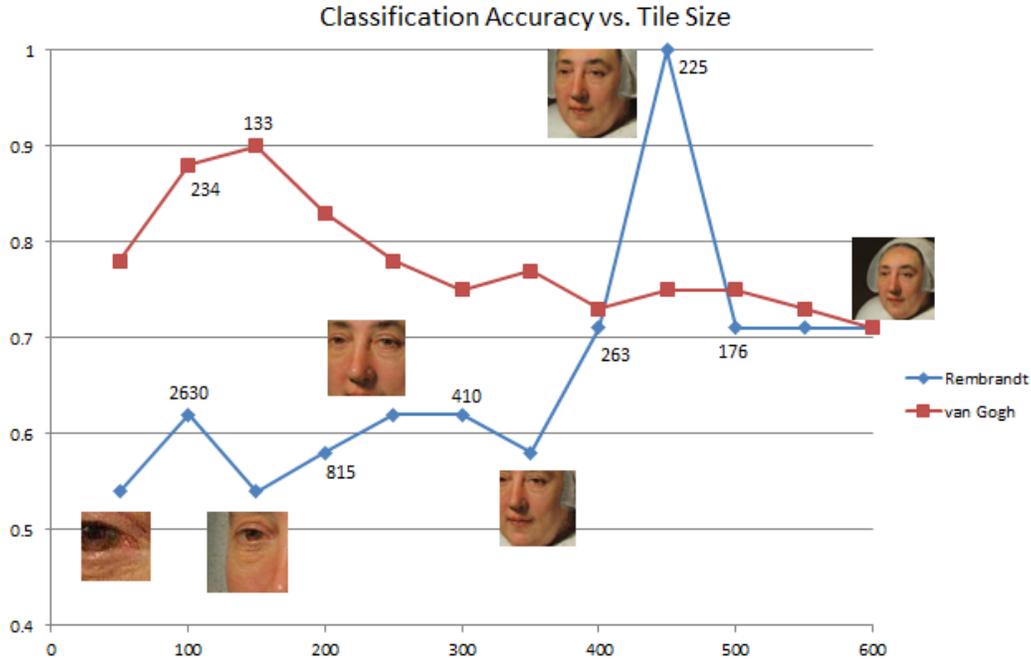

Fig. 4: Classification accuracy vs. tile size. Representative features from the same Rembrandt portrait and average number of entropy-filtered tiles per image are shown for various tile sizes

We found that one strategy for improving accuracy is to augment the overlap among testing tiles and, critically for larger tiles, in the training set as well to boost numbers. Even extreme levels of overlap can significantly increase accuracy as a result of the larger tile population despite data redundancy. At 50% overlap, we obtained about 1000 Rembrandt training tiles of size 450×450. Increasing the overlap to 92% resulted in over 15,000 training tiles and improved the classification accuracy from 83% to 100%. We followed a different augmentation strategy for van Gogh tiles, as discussed below, but evaluation of the final results revealed that many of the misclassified van Gogh paintings had prediction probabilities close to the decision boundary. Averaging the painting-level probabilities produced by two models or, better, using linear optimization to weight each model's output boosts overall van Gogh accuracy from 90% to 92%. In particular, we assign weights so as to minimize the average image classification error $E$, which we define as an L1 norm:

$$E = \frac{1}{n}\sum_{i=1}^{n}|0.5 - \hat{y}_i|$$

where $n$ is the number of misclassified paintings, 0.5 is the decision boundary, and $\hat{y}_i$ is the (erroneous) predicted probability for the $i^{\text{th}}$ painting. Combining results in this way seems reasonable because, first, multi-scale CNN training has been applied successfully to artwork classification (van Noord et al., 2017), and second, any professional attribution exercise will consider a work at different feature scales.

OVERALL COMPARISON

| Model | Rembrandt | van Gogh |
|---|---|---|
| *5-layer* | | |
| Best accuracy/tile size | 100% / 450 | 90% / 100 |
| Best accuracy variance | 0.10 | 0.05 |
| Best accuracy/tile combination | — | 92% / 100+150 |
| Randomly selected tiles | 71% / 450 | 80% / 100 |
| Rectangular tiles | (test not performed) | 84% / 150 |
| | | |
| *Entropy criterion relaxed 1%* | | |
| Best accuracy/tile size | (test not performed) | 92% / 150 |
| Best accuracy/tile combination | — | 94% / 100+150 |
| | | |
| *ResNet50* | | |
| Best accuracy/tile size | (Failed) | 61% / 150 |
| | | |
| *VGG16* | | |
| Best accuracy/tile size | (Failed) | 53% / 150 |

Table 1: Performance comparison among models and methodologies for tile definition and selection. For van Gogh we obtained more qualifying tiles, and better accuracy, by relaxing the entropy criterion by 1%. Variance, which represents prediction consistency, was computed across tile-level probabilities for each painting in the test set; the reported value is the average variance over all paintings in the test set.

We compared our entropy-based selection criterion with random tile selection. As shown in Table 1, we found that the entropy criterion produces a substantial increase in accuracy for Rembrandt and an appreciable but smaller benefit for van Gogh. To investigate the difference, we obtained and plotted, for both artists and the comparative images, entropies associated with optimally sized tiles. The results appear in Fig. 5. We see that, for van Gogh, a random tile is likely to have high entropy, so the performance of randomly selected tiles inherently approaches that of tiles selected explicitly based on image entropy.

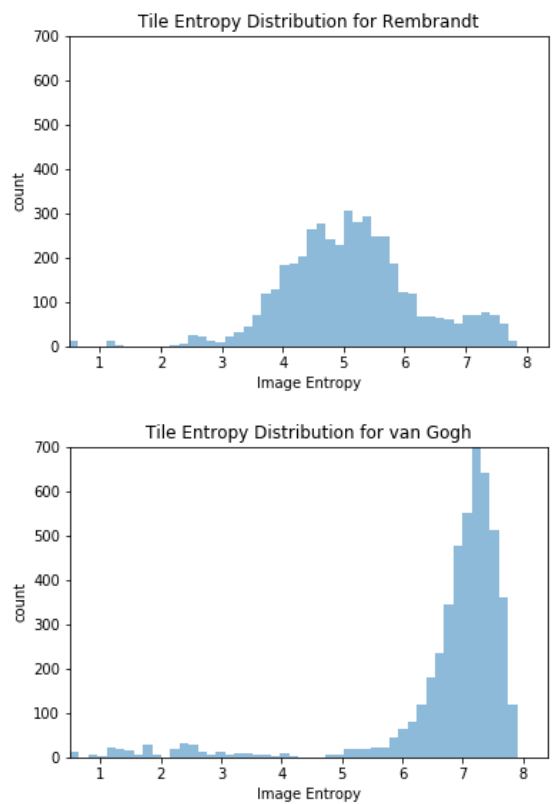

Fig. 5: Image entropy distributions for Rembrandt
(450×450) and van Gogh (100×100) tiles

The skewed distribution for van Gogh tiles also accounts for the much lower tile-retention rate noted above. While tile entropy peaks at the average painting entropy of 7.43, relatively few tiles have entropies larger than that. By relaxing the discrimination criterion by 1% — that is, retaining tiles with entropies equal to or exceeding 99% of the painting entropy — we shift the entropy cutoff toward the mean tile-level image entropy, with the result that the tile population nearly doubles (from 5260 to 9402 tiles of size 150×150, for example). Retraining and retesting with van Gogh tiles selected with this relaxed criterion, we found that accuracy increased from 90% to 94% with weighted probabilities (though the average classification error was comparable). Further augmentation by increasing tile overlap did not improve accuracy despite the significant increase in tile numbers.

We performed cross-validation studies at the painting level using datasets sized as noted previously. Table 2 lists the average number of tiles per image, for training and testing purposes, that qualified for evaluation prior to data augmentation as described above, as well as the number of tiles actually used per fold following augmentation. To maintain a reasonably sized test set of 19 images for Rembrandt (out of a total of 76 images), we used four-fold cross-validation; the larger van Gogh dataset easily accommodated five-fold cross-validation with 30 images per fold.

| Artist | Tile Size | Avg. Tiles/Image and Tiles/Fold | Cross-Validation Range |
|--------|-----------|----------------------------------|------------------------|
| *Rembrandt* | 450×450 | 29 / 4280 | 0.88–1.0 |
| *van Gogh* | 150×150 | 146 / 4030 | 0.88−0.95 |
| *van Gogh* | 100×100 | 360 / 7112 | 0.88−0.91 |

Table 2: For the best tile sizes, we performed four-fold (for Rembrandt) and five-fold (for van Gogh, with the relaxed entropy criterion) cross-validation

Overfitting invariably set in short of 30 epochs. At the best-performing 450×450 tile size, for example, accuracy for Rembrandt peaked after 26 epochs and subsequently declined, as shown in Fig. 6.

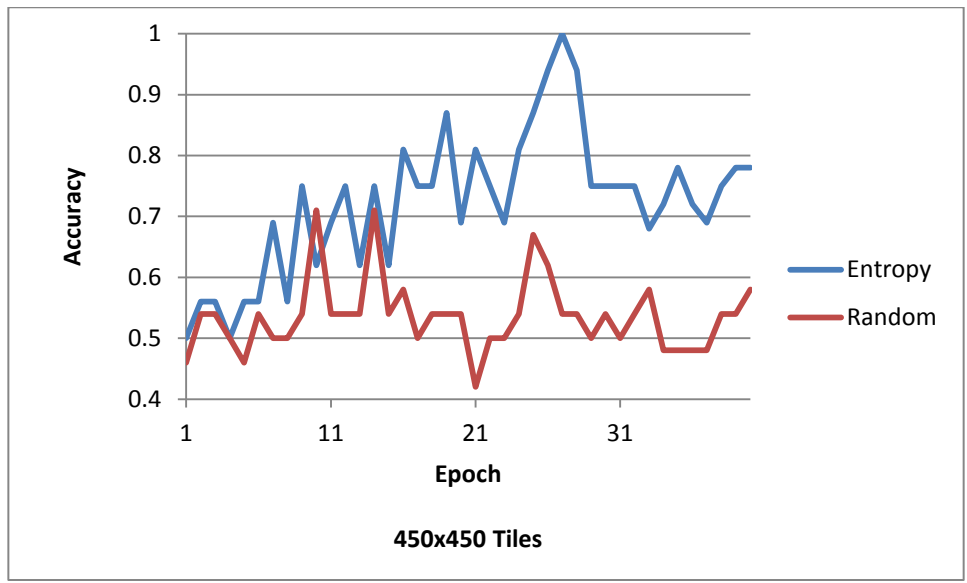

Fig. 6: Test accuracy as a function of training epoch for 450×450 Rembrandt tiles

Because of the ensemble averaging, validation accuracy at a tile level bears little relation to overall classification accuracy, so early stopping is not a useful strategy, nor are validation metrics particularly meaningful. Training accuracy generally exceeds 85% after the first epoch and 99% after five epochs; it too bears little relation to test accuracy. Consquently, we save the model after each epoch and test against all saved versions following training to identify the one that performs best.

Fig. 6 also illustrates training behavior using randomly selected tiles, which do not produce a distinct accuracy peak — supporting our contention that high image entropy enhances CNN performance for this classification task.

We put our approach to the test by evaluating 10 works that have been subject to attribution controversy, and Table 3 compares our results to the current consensus of scholars. Also appearing in Table 3 is our successful classification of a van Gogh fake painted by a famous forger, John Myatt.

| Title | Scholarly Consensus | Our Classification | Probability |
|---|---|---|---|
| *Man with the Golden Helmet* | School of Rembrandt | Rembrandt | 0.85 |
| *Portrait of A Young Gentleman* | Rembrandt | Rembrandt | 0.81 |
| *Portrait of Elisabeth Bas* | Ferdinand Bol | Not Rembrandt | 0.39 |
| *The Polish Rider* | Rembrandt | Rembrandt | 0.57 |
| *Portrait of a Man ("The Auctioneer")* | Follower of Rembrandt | Not Rembrandt | 0.29 |
| *Portrait of A Young Woman (Kress Collection/Allentown Art Museum)* | Rembrandt | Rembrandt | 0.82 |
| *View of Auvers-sur-Oise* | van Gogh | van Gogh | 0.90 |
| *Garden of the Asylum* | van Gogh | van Gogh | 0.98 |
| *Three Pairs of Shoes* | van Gogh | van Gogh | 0.54 |
| *Church at Auvers* | (John Myatt fake) | Not van Gogh | 0.29 |

Table 3:  Overall classifications and average probabilities across tested tiles for controversial works whose attributions have shifted over time, and for a known fake (i.e., a work painted to resemble van Gogh but not passed off as his work)

The Rembrandt controversies, some of which have raged for over a century, are well described in van de Wetering, 2017.  Bailey, 1997; Bailey, 2005; Hendriks et al., 2001; and Hulsker, 1996 discuss questions surrounding the listed van Gogh works.  John Myatt's career turn from notorious forger to celebrity painter of "genuine fakes" is chronicled in Charney, 2015.

Our best models, in isolation or combined in a weighted average with a second model, classify all but one work in accord with expert consensus.  (Table 3 reflects weighted-average probabilities.)  The exception is *Man with the Golden Helmet*, which our algorithm decisively classifies as a Rembrandt.  Probing more deeply into the scholarship surrounding the Rembrandt Research Project's de-attribution of this painting, we find this justification: "In particular the thick application of paint to the helmet in contrast to the conspicuously flat rendering of the face, robe and background, which are placed adjacent to each other without a transition, does not correspond to Rembrandt's way of working" (Kleinert et al.). Given the relative inaccuracies of Rembrandt models based on detail-level tiles, we modestly question the reliability of attribution judgments based on fine surface features.

## V.    LIMITATIONS, STRATEGIES, EXTENSIONS

We were encouraged by a prediction accuracy exceeding 90% using relatively small training sets drawn from a small number of images.  Thanks to the effects of ensemble averaging, we were able to examine candidate works computationally at multiple feature scales and average out most of the classification errors.  Our approach avoids hand-engineered features but requires careful attention both to data curation and tile sizes.  Indeed, even models trained on large and diverse datasets can never fully settle issues of authenticity, because no training set can include all past or future imitative efforts.  Trained models will always be incomplete and potentially spoofed by forgeries outside the generalization achieved by training.

Although it is straightforward to identify an optimal tile size and combine results from multiple sizes with weighting, it is important not to allow the mechanics to obscure the classification context.  To analyze paintings, candidate tile sizes should range from brushstroke detail to compositional motifs.  The size range for medical images, by contrast, will be dictated by considerations of anatomy and pathology. Researchers developing a deep-learning system for classifying biopsy images, for example, recently discovered that the most predictive features for breast-cancer survival lie in the region surrounding tumor cells — not the cells themselves (Dolgin, 2018).  Too small and skewed a size range (selected to resolve cells, for example) would miss these essential larger-scale features.

More broadly, we see the Salient Slices technique as useful for analyzing and classifying images having heterogeneous visual features and where visual information diversity correlates with classification accuracy. Representational art strikes us as one such domain. Abstract art, by contrast — particularly where the trace of the artist's hand is not necessarily manifest in high-entropy image regions — would not prove as good a candidate for this technique. A compelling feature of Jackson Pollock's drip paintings, for example, is the way the center is unified and organized by diminishing content toward the edges. If sparse edges are key to identifying Pollocks, our technique would perform poorly — perhaps so poorly as to be useful, if the selection criterion were flipped from high to low image entropy.

We hope that Salient Slices will prove effective as a tool to sharpen the expert's eye; we know it will never be a replacement.

***End Notes***

---